\documentclass{bmvc2k}

\usepackage{enumitem}
\usepackage{multirow}
\usepackage{multicol}
\usepackage{adjustbox}
\usepackage{bbm}
\usepackage{amsmath,amsfonts,amssymb}
\newcommand{\Lagr}{\mathcal{L}}
\newcommand{\Soln}{DEX}
\newcommand{\SolnLite}{DEXLite}

\title{DEX: Domain Embedding Expansion for Generalized Person Re-identification}

\addauthor{Eugene P.W. Ang}{phuaywee001@e.ntu.edu.sg}{1,2}
\addauthor{Lin Shan}{shan.lin@ntu.edu.sg}{1}
\addauthor{Alex C. Kot}{eackot@ntu.edu.sg}{1}

\addinstitution{
 Rapid-Rich Object Search (ROSE) Lab\\
 Nanyang Technological University\\
 Singapore
}
\addinstitution{
 Defence Science and Technology Agency\\
 Singapore
}

\runninghead{Ang, Lin, Kot}{Domain Embedding Expansion}

\def\eg{\emph{e.g}\bmvaOneDot}

\begin{document}
\maketitle
\begin{abstract}

In recent years, supervised Person Re-identification (Person ReID) approaches have demonstrated excellent performance. However, when these methods are applied to inputs from a different camera network, they typically suffer from significant performance degradation. Different from most domain adaptation (DA) approaches addressing this issue, we focus on developing a domain generalization (DG) Person ReID model that can be deployed without additional fine-tuning or adaptation. In this paper, we propose the Domain Embedding Expansion (\Soln{}) module. \Soln{} dynamically manipulates and augments deep features based on person and domain labels during training, significantly improving the generalization capability and robustness of Person ReID models to unseen domains. We also developed a light version of \Soln{} (\SolnLite{}), applying negative sampling techniques to scale to larger datasets and reduce memory usage for multi-branch networks. Our proposed \Soln{} and \SolnLite{} can be combined with many existing methods, Bag-of-Tricks (BagTricks), the Multi-Granularity Network (MGN), and Part-Based Convolutional Baseline (PCB), in a plug-and-play manner. With \Soln{} and \SolnLite{}, existing methods can gain significant improvements when tested on other unseen datasets, thereby demonstrating the general applicability of our method. Our solution outperforms the state-of-the-art DG Person ReID methods in all large-scale benchmarks as well as in most the small-scale benchmarks.



\end{abstract}

\section{Introduction}
\label{sec:intro}


Person Re-identification (Person ReID) has achieved impressive performance on academic benchmarks in recent years. However, generalization issues prevent its transition into the applied world. For example, a model trained on one Person ReID domain using standard techniques can achieve high accuracy when independently tested within the same domain, but its performance degrades drastically when tested on a unseen domain. This reveals a lack of generalization ability in single-dataset supervised models and suggests that the features learned by these models over-fit the training domain instead of capturing the general features relevant for person discrimination. In recent years, much of Person ReID research has focused on unsupervised domain adaptation (DA)~\cite{MMFA,TJ-AIDL,UMDL,SPGAN}, which uses unlabeled data collected from the target domain to alleviate the domain over-fitting problem. However, in many real-world applications, access to the data from the target domain beforehand may not be a valid assumption to make. Therefore, domain generalization (DG) is a more practical strategy for that problem. DG methods leverage the different distributions of multiple datasets to reduce domain bias.

\begin{figure}[t]
\includegraphics[width=12.6cm]{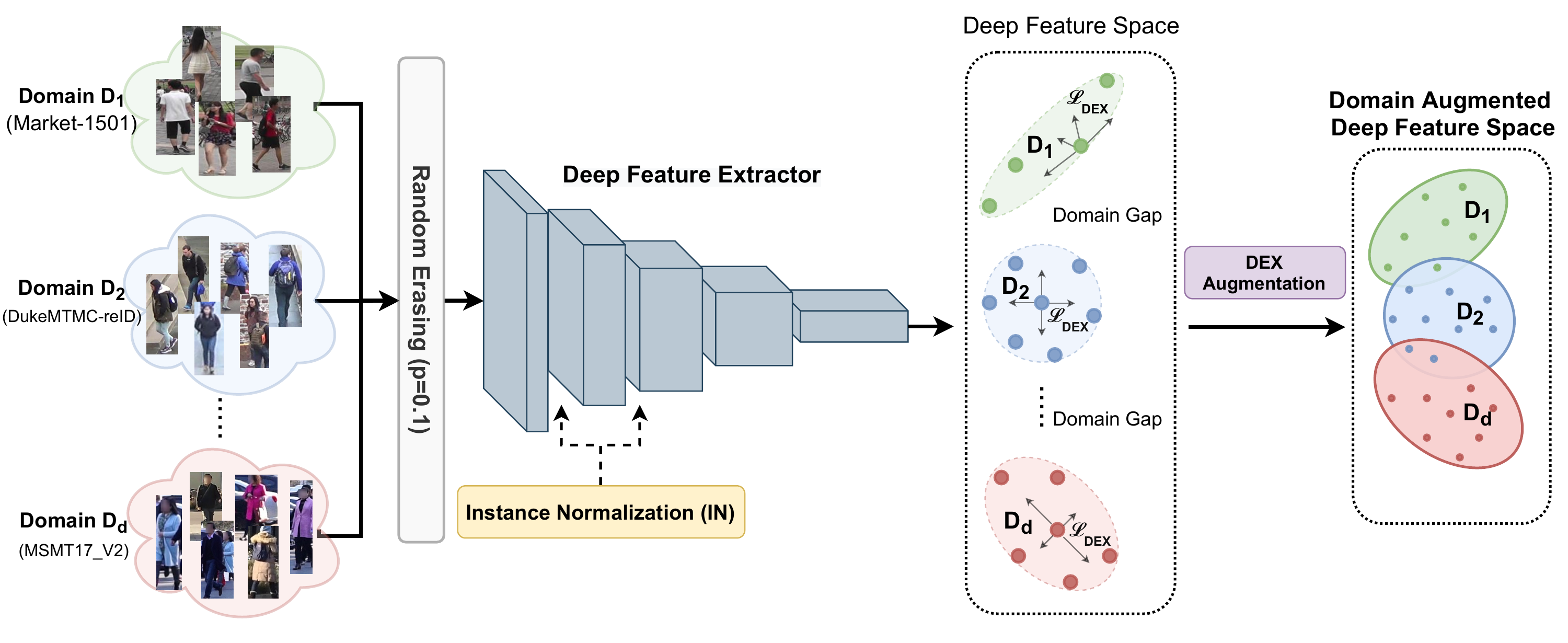}
\caption{Our proposed solution illustrated. Our baseline is enhanced by reducing the probability of RE and applying IN at early layers of a deep feature extractor. In feature space, our deep augmentation method \Soln{} improves domain manifold coverage by implicitly projecting the feature points in directions of the domain distribution. Best viewed in color. }
\label{fig:overall-method}
\end{figure}

Current DG Person ReID methods involve complex frameworks such as meta-learning, hyper-networks and memory banks~\cite{DIMN,QAConv,M3L} supported by custom loss functions and normalizations. In this paper, we tackle the DG problem of Person ReID from a novel perspective using deep feature augmentation. We propose \textbf{D}omain \textbf{E}mbedding E\textbf{x}pansion (\Soln{}), a deep feature augmentation module that leverages person and domain labels to fill the domain gap between deep features during training. Many Generative Adversarial Network (GAN)~\cite{Goodfellow2014GenerativeNets} based methods also transfer style information from one domain to another while preserving person identity features. However, such methods are computationally expensive and require a  nontrivial GAN-training stage. Our proposed \Soln{} implicitly projects extracted deep features across domain manifolds during the training process, as shown in Figure \ref{fig:overall-method}. Applying \Soln{} on our baseline (DualNorm~\cite{DualNorm} + BagTricks~\cite{StrongBaseline} over a ResNet-50~\cite{ResNet} backbone) allows us to outperform state-of-the-art methods on all large-scale benchmarks by a wide margin. Integrating \Soln{} on other popular Person ReID architectures such as Multi-Granularity Network (MGN)~\cite{MGN} and Part-Based Convolutional Baseline (PCB)~\cite{PCB}, demonstrates consistent improvement in model generalization. When utilizing multiple Person ReID datasets for training, the increase in unique person-identities (PIDs) could lead to a huge demand for GPU memory. We also developed a memory-light version of \Soln{} (\SolnLite{}) that applies negative sampling to reduce computation, memory consumption, and training time. \SolnLite{} supports datasets with large numbers of PIDs, larger batch sizes, and multi-branch model architectures.

To summarize, our contributions are as follows:

\begin{itemize} [noitemsep,nolistsep]
  \item We propose \Soln{}, a deep feature data augmentation method tailored for the multi-domain generalization Person ReID problem that leverages domain labels to implicitly project deep features over domain manifolds.
  \item With \Soln{} and Instance Normalization (IN), a simple ResNet50 backbone can outperform state-of-the-art performance on all large scale Person ReID DG benchmarks (Market-1501, DukeMTMC-reID, CUHK03 and MSMT17) by a wide margin. 
  \item For memory-limited machines, we developed \SolnLite{}, a memory-light version of \Soln{}, that uses negative sampling to reduce memory use during training, enabling the use of this technique on a broader range of problems. We apply \SolnLite{} on two multi-branch architectures, MGN and PCB, to demonstrate significant improvements. 
  \item \SolnLite{} also outperforms state-of-the-art methods in most of the small-dataset DG benchmark metrics and closely matches the top performers for the remaining metrics.
  \item Our proposed method is straightforward and does not require complex frameworks or specialized neural network architectures.
\end{itemize}



\section{Related Work}
\label{sec:related-work}

\noindent \textbf{Person Re-identification.} 
Deep learning based Person ReID approaches developed in recent years, such as PCB~\cite{PCB}, BagTricks~\cite{StrongBaseline}, and MGN \cite{MGN} have achieved impressive accuracy in Person ReID. These methods are usually trained and evaluated within the same dataset. However, due to the difficulty of data collection and limited numbers of cameras and pedestrians, existing datasets suffer from limited variability in location, weather, pedestrian clothing, illumination, and camera color settings. Such limitations induce a strong domain bias during training and significantly degrade model performance when testing on other domains. Hence, many domain adaptation (DA) methods such as UMDL~\cite{UMDL}, SPGAN~\cite{SPGAN}, TJ-AIDL~\cite{TJ-AIDL}, MMFA~\cite{MMFA} were proposed to bridge the gap between domains for the Person ReID problem. DA methods utilize unlabeled data from the target domain to generate pseudo-labels or transfer the target domain image style to the source domain. Although many DA approaches yield good performance improvements when dealing with the cross-domain problem, these methods still require a large amount of data from the target domain, hence limiting their application to real-world Person ReID problems. Domain generalization (DG) methods, on the other hand, operate under a more challenging scenario, assuming no access to any target domain data. The common objective among DG methods is to learn a general universal feature representation that is robust to domain shift. DualNorm \cite{DualNorm} first introduced instance normalization (IN) in the early stages of the network to normalize the style and content variations of the datasets. On the other hand, MMFA-AAE~\cite{MMFA-AAE} used a domain adversarial learning approach to remove domain-specific features. Latest DG methods such as DIMN~\cite{DIMN}, QAConv~\cite{QAConv} and M$^3$L~\cite{M3L} used hyper-networks or meta-learning coupled with a memory bank strategy. These meta-learning approaches require complicated training procedures, which makes model optimization difficult. Different from all existing methods, \Soln{} addresses the domain generalization issue from a novel deep feature domain augmentation perspective. Our method uses widely available tools, a ResNet-50 backbone, and does not require complex frameworks or training procedures.

\noindent\textbf{Augmentation.} Most of the proposed augmentation solutions in Person ReID use GANs ~\cite{Goodfellow2014GenerativeNets} to transfer target domain image styles to source domain images. LSRO~\cite{DukeMTMC-reID} and DG-Net~\cite{DG-Net} first used GANs as an augmentation for better representation learning in Person ReID. SPGAN~\cite{SPGAN} and PTGAN~\cite{MSMT17} then expanded this to cross dataset scenarios. However, these GAN-based approaches mainly transfer image-style differences across different domains without considering other semantic differences such as clothing styles, weather, etc. Furthermore, training such generative models is difficult and computationally expensive, with the final outputs exhibiting noticeable artifacts. Our method focuses on a new augmentation direction for Person ReID, which directly performs semantic transformations in the deep feature space. Many works~\cite{DFI,DeepAugment,ISDA} discovered that it was possible to meaningfully alter the semantics of image samples (\eg{} changing the color of an object) by perturbing their corresponding deep features in specific directions. DeepAugment~\cite{DeepAugment} proposed to pass images through a pretrained image-to-image model while altering the features via stochastic operations in the model, producing multiple diverse yet semantically consistent images for training. However, training good image generation models is challenging and time-consuming, and the increase in data leads to increased memory use and training time. ISDA~\cite{ISDA} proposed to project deep features by sampling semantic directions from class-conditional covariance matrices, implicitly performing the projections and reducing memory footprint by optimizing a surrogate robust upper-bound loss. Our proposed \Soln{} is tailored for the multi-domain generalization Person ReID problem. It expands feature samples in meaningful domain-semantic directions while preserving person identity information.

\section{Method}
\label{sec:method}
\vspace{-1em}
\subsection{Domain Embedding Expansion (\Soln{})} 

\begin{figure}[t]
\centering
\includegraphics[width=12.4cm]{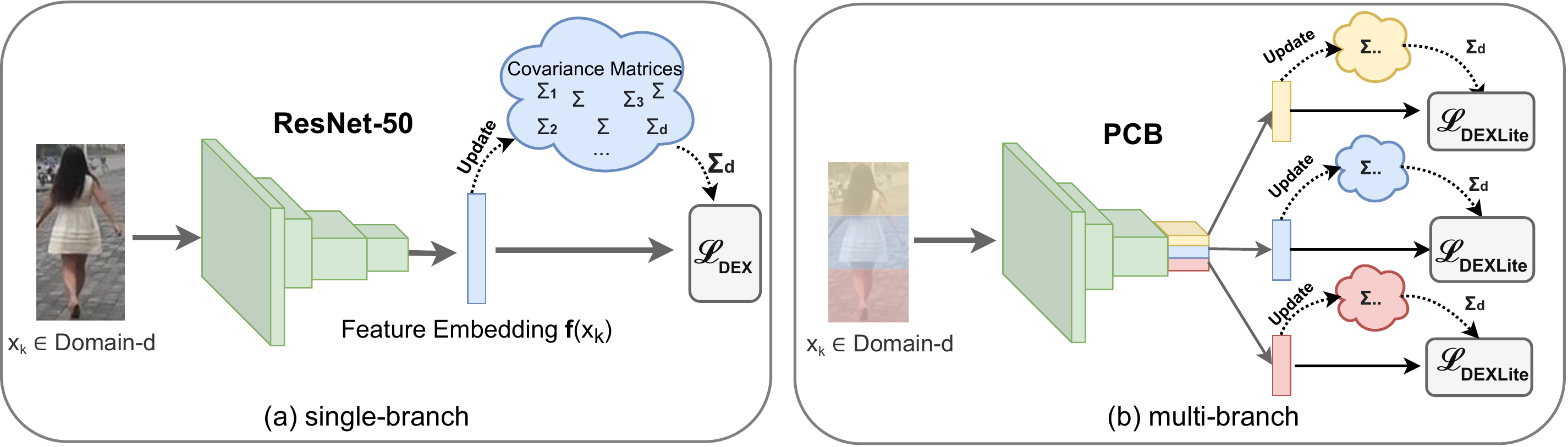}
\caption{\Soln{}: (a) This illustrates how \Soln{} is applied to a single classifier branch; (b) Given a model with multiple classifier branches, such as PCB shown above, we store and update a separate set of covariance matrices for each branch and also compute a per-branch \SolnLite{} loss. For illustration only, we reduced the PCB network to 3 stripes. }
\label{fig:our-method}
\end{figure}

To improve feature representation learning and close domain gaps, we propose \Soln{}, a deep feature augmentation technique specially adapted for multi-domain generalization problems. Building on previous deep feature augmentation methods~\cite{DFI,DeepAugment,ISDA}, we perturb deep features generated by the model as a form of augmentation. Each deep feature is projected along directions sampled from a zero-mean normal distribution with covariance matrix estimated from the feature's domain; this is so that features from different domains are projected in directions that correspond to semantic transformations meaningful to their domain. The covariance matrices are estimated online during training. Similar to ISDA~\cite{ISDA}, the data samples are not explicitly perturbed. Instead, we optimize a proxy loss function that upper-bounds the expected cross-entropy loss of perturbed data samples. Figure \ref{fig:our-method} presents an overview of our method.

\noindent\textbf{Formulation} \Soln{} is a modification to the classification branch of the model. The regular softmax loss $\Lagr_{soft}$ is computed from the feature output of the model and the weights of its fully connected layer as such: 
\begin{equation} 
\Lagr_{soft}=\frac{1}{N} \sum_{i=1}^{N} -\log \left( \frac{\exp(\mathbf{w}_{y_i}^{\intercal} \mathbf{a_i} )}{\sum_{j=1}^{C} \exp(\mathbf{w}_{j}^{\intercal} \mathbf{a_i})} \right),
\label{eqn:softmax}
\end{equation}
where $N$ is the batch size indexed by $i$, $C$ is the number of unique PIDs indexed by $j$, $\mathbf{a_i}$ are deep features output by the model and $\mathbf{w}$ are weight vectors of the model's fully connected layer; biases are omitted in our classifier layer. The design rationale behind \Soln{} is derived from \cite{ISDA}. First we consider the \emph{expected} softmax loss if the deep features were projected along domain-conditional covariance directions $\Sigma_{d_i}$: 
\begin{equation} 
\begin{split}
\Lagr_{\infty} & = \frac{1}{N} \sum_{i=1}^{N} \mathbb{E}_{ \mathbf{\Tilde{a}_i}  } \left[ -\log \left( \frac{\exp(\mathbf{w}_{y_i}^{\intercal} \mathbf{ \Tilde{a}_i } )}{\sum_{j=1}^{C} \exp(\mathbf{w}_{j}^{\intercal} \mathbf{ \Tilde{a}_i  })} \right) \right] \\
& = \frac{1}{N} \sum_{i=1}^{N} \mathbb{E}_{ \mathbf{\Tilde{a}_i}  } \left[ \log \left(  
\sum_{j=1}^{C} \exp( (\mathbf{w}_{j}^{\intercal} - \mathbf{w}_{y_i}^{\intercal}) \mathbf{ \Tilde{a}_i  }) \right) \right],
\end{split}
\label{eqn:loss-infinity}
\end{equation}
\noindent where $ \mathbf{ \Tilde{a}_i } \sim \mathcal{N}( \mathbf{a_i} , \lambda \Sigma_{d_i}) $ are the augmented features assumed to be normally distributed around $\mathbf{a_i}$ with domain-conditional covariance $\Sigma_{d_i}$, and $\lambda \ge 0$ controls the strength of the augmentation. Applying Jensen's Inequality, $\mathbb{E}[\log (X)] \le \log (\mathbb{E} [X])$, we can move the $\log$ out of the expectation to get:

\begin{equation} 
\Lagr_{\infty} \le \frac{1}{N} \sum_{i=1}^{N} \log \left( \sum_{j=1}^{C} \mathbb{E}_{ \mathbf{\Tilde{a}_i}  } [ \exp( (\mathbf{w}_{j}^{\intercal} - \mathbf{w}_{y_i}^{\intercal}) \mathbf{ \Tilde{a}_i  })  ] \right)
\label{eqn:loss-jensen}
\end{equation}

We apply the moment generating function $\mathbb{E}[\exp(tX)]=\exp(t \mu + \frac{1}{2} \sigma^2 t^2)$, $X \sim \mathcal{N}(\mu,\sigma^2)$, substituting $t$ with $(\mathbf{w}_{j}^{\intercal} - \mathbf{w}_{y_i}^{\intercal})$ and $X \sim \mathcal{N}(\mu, \sigma^2)$ with $\mathbf{ \Tilde{a}_i } \sim \mathcal{N}( \mathbf{a_i} , \lambda \Sigma_{d_i})$, to derive our loss: 
\begin{equation}
\Lagr_{\Soln{}}(\lambda) =\frac{1}{N} \sum_{i=1}^{N} -\log \left( \frac{\exp(\mathbf{w}_{y_i}^{\intercal} \mathbf{a_i} )}{\sum\limits_{j=1}^{C} \exp(\mathbf{w}_{j}^{\intercal} \mathbf{a_i} + \frac{\lambda}{2}(\mathbf{w}_{j}^{\intercal}-\mathbf{w}_{y_i}^{\intercal})\Sigma_{d_i} (\mathbf{w}_{j}-\mathbf{w}_{y_i}) )} \right) \ge \Lagr_{\infty}
\label{eqn:isda-dom}
\end{equation}
Thus, $\Lagr_{\Soln{}}$ upper bounds the \emph{expected} softmax loss of projecting deep features $\mathbf{ a_i }$ over directions encoded in $\Sigma_{d_i}$, and we can directly optimize $\Lagr_{\Soln{}}$ to reap the benefits of augmentation while avoiding the extra computation of explicitly projecting features.

Different from \cite{ISDA} that performs implicit augmentation in class-semantic directions, \Soln{} explores meaningful \emph{domain-semantic directions} in the deep manifold space for different Person ReID datasets. ISDA was proposed in the context of image classification problems and is impractical for multi-domain Person ReID problems where thousands of per-class covariance matrices collectively impose a prohibitive memory overhead. Approximations have been proposed by \cite{ISDA} to overcome this memory issue, but even so, our experiments show that applying ISDA naively fails to improve the baseline for all benchmarks. Our analysis reveals that Person ReID datasets have very few samples-per-class (10-25 on average) which makes covariance estimation unreliable, compared to the datasets used in ISDA which have 500-5,000 on average. \Soln{} is a superior design for two reasons. Firstly, there are much fewer domains than classes so we need not rely on approximations to reduce memory use. Secondly, since the number of samples-per-domain is very large (numbering in the 10,000s), the domain-conditional covariance estimates are stable. The wide domain gaps observed in DG Person ReID show that \emph{deep features invariably encode domain semantics}; consequently, their estimated domain-conditional covariance matrices would contain meaningful domain-semantic directions. Experimentally, \Soln{} significantly improves performance for every benchmark. Ablation studies comparing ISDA with our method are presented in Section \ref{sec:experiments}, and a detailed study of the differences between Person ReID datasets and those studied in ISDA are presented in Supplemental Material.

\subsection{Domain Embedding Expansion Lite (\SolnLite{})}
\Soln{} requires additional memory during training. During each back-propagation step, intermediate tensors require {\em extra} space on the order of $O(B_{CE} (Df^2 + CN))$. $B_{CE}$ is the number of cross-entropy loss branches that apply \Soln{}, $D$ is the number of domains, $f$ is the feature dimension, $C$ is the number of PIDs in the training set with batch size $N$. By using approximations~\cite{ISDA}, we can reduce the complexity to $O(B_{CE} (Df + CN))$. However, there are many reasonable scenarios where $B_{CE}$ and $C$ can dominate the complexity. If there are multiple classification loss branches as in the case of PCB~\cite{PCB} and MGN~\cite{MGN}, or if the number of PIDs explodes as a result of merging more source domains, memory overhead grows infeasible. For these use cases, we developed a lightweight version of \Soln{} (\SolnLite{}) that applies negative sampling to alleviate this issue. Negative sampling is most popularly used in classification problems with a large number of classes, most notably in training language models to speed up softmax loss computation over a large vocabulary~\cite{Word2Vec}. Computing the full denominator of Equation \ref{eqn:isda-dom} is memory intensive and wasteful since most classes are negatives; in our sampling strategy, a batch size of $B$ only has $\frac{B}{4}$ unique identities, meaning that if we train on a dataset with 10,000 PIDs, a batch size of 32 contains 8 unique positive IDs that account for only 0.08\% of all identities. We propose to sample only a subset of negatives to reduce the total number of classes considered during \Soln{} augmentation. The loss function of Equation \ref{eqn:isda-dom} then changes to:
\begin{equation}
\Lagr_{\SolnLite{}}(\lambda)=\frac{1}{N} \sum_{i=1}^{N} -\log \left( \frac{\exp(\mathbf{w}_{y_i}^{\intercal} \mathbf{a_i} )}{\sum_{j \in P_s} \exp(\mathbf{w}_{j}^{\intercal} \mathbf{a_i} + \frac{\lambda}{2}(\mathbf{w}_{j}^{\intercal}-\mathbf{w}_{y_i}^{\intercal})\Sigma_{d_i} (\mathbf{w}_{j}-\mathbf{w}_{y_i}) )} \right),
\label{eqn:isda-dom-lite-full}
\end{equation}
where $P_s$ is the set of sampled PIDs, including positives; the space complexity is reduced to $O(B_{CE} (Df + |P_{s}|N))$, with $|P_{s}| \ll C$. The development of \SolnLite{} enables us to apply this augmentation onto a wider range of models and datasets with many PIDs. As an added bonus, it also speeds up training. To demonstrate its capability, we implement \SolnLite{} on MGN and PCB and report significant improvements in Section \ref{sec:experiments}.

\subsection{Overall Loss Function}
Our proposed method is based on simple ResNet-50 backbone and trained with three losses: softmax loss over PIDs $\Lagr_{soft}$, triplet loss $\Lagr_{tri}$~\cite{TriNet} and center loss $\Lagr_{cen}$~\cite{Wen2016ARecognition}. $\Lagr_{soft}$ is defined in Equation \ref{eqn:softmax} and definitions of $\Lagr_{tri}$ and $\Lagr_{cen}$ are provided in Supplemental Material. We train the baseline with combined loss $\Lagr_{base}$ with $\beta_{soft}=1.0$, $\beta_{tri}=1.0$, and $\beta_{cen}=5\times10^{-4}$:

\begin{equation}
\Lagr_{base} = \beta_{soft} \Lagr_{soft} + \beta_{tri} \Lagr_{tri} + \beta_{cen} \Lagr_{cen}
\end{equation}

The overall loss $\Lagr_{overall}(t)$ is parameterized by current epoch $t$ which controls the strength of the implicit augmentation. $\Lagr_{\Soln{}}$ and $\Lagr_{\SolnLite{}}$ are defined in Equation \ref{eqn:isda-dom} and Equation \ref{eqn:isda-dom-lite-full}. At epoch $t$, $\lambda_{t}=\frac{t-1}{T-1}\lambda$ with $\lambda=7.5$; we gradually increase $\lambda_{t}$ to pay more attention to later model features as they become more informative over time. For baseline and \Soln{} we train for 60 epochs ($T=60$) with a batch size of 32 using the Adam~\cite{Kingma2015Adam:Optimization} optimizer with a learning rate schedule similar to ~\cite{StrongBaseline}. Full details are available in Supplemental Material.
\begin{equation}
\Lagr_{overall}(t) = \Lagr_{\Soln{}/\SolnLite{}}(\lambda_{t}) + \beta_{tri} \Lagr_{tri} + \beta_{cen} \Lagr_{cen}
\end{equation}

\section{Experiments}
\label{sec:experiments}
\subsection{Datasets and Settings}
We model our experiments after the most recent state-of-the-art method M$^3$L \cite{M3L} in DG Person ReID in four large-scale benchmarks: Market-1501 \cite{Market-1501}, DukeMTMC-reID \cite{DukeMTMC-reID}, CUHK03 \cite{CUHK03} (or its new partition CUHK03-NP \cite{Re-ranking}) and MSMT17\_V2 \cite{MSMT17}. Table \ref{tab:datasets} breaks down the number of IDs and images in each of the training, query and gallery splits for each dataset. For simplicity and clarity, we denote Market-1501, MSMT17, DukeMTMC-reID, CUHK03, and CUHK03-NP as M, MS, D, C, and C-NP.
\begin{table}[h]
\centering
\begin{adjustbox}{width=\textwidth}
\begin{tabular}{l|c|c|c|c|c|c|c}
\hline
Dataset       & Abbreviation & Train-IDs & Train-Images & Query-IDs & Query-Images & Gallery-IDs & Gallery-Images \\ \hline
Market-1501~\cite{Market-1501}   & M            & 751       & 12,936       & 750       & 3,368        & 750         & 15,913         \\
MSMT17\_V2~\cite{MSMT17}    & MS           & 1,041     & 32,621       & 3,060     & 11,659       & 3,060       & 82,161         \\
DukeMTMC-reID~\cite{DukeMTMC-reID} & D            & 702       & 16,522       & 702       & 2,228        & 1,110       & 17,661         \\
CUHK03-NP~\cite{Re-ranking}     & C-NP         & 767       & 7,365        & 700       & 1,400        & 700         & 5,332          \\
CUHK03~\cite{CUHK03}        & C            & 1,367     & 26,263       & -         & -            & -           & -              \\ \hline
\end{tabular}
\end{adjustbox}
\caption{Dataset details. For testing we always use the query/gallery split of C-NP. We follow ~\cite{M3L} to select either C or C-NP as source domain. C is never used for testing.}
\label{tab:datasets}
\end{table}
\vspace{-1em}


\subsection{Comparison with State-of-the-art Methods under New Evaluation}
Following the new evaluation methodology proposed in \cite{M3L}, we use the detected test subset of the CUHK03 new protocol, CUHK-NP (detected), for testing and CUHK03 as one of the source domains for training. The training splits of any three datasets are combined into a training set and the query/gallery split of the remaining dataset is used for testing.

\begin{table}[h]
\centering
\begin{adjustbox}{width=\textwidth}
\begin{tabular}{c|l|c|c|c|l|cc}
\hline
\multirow{2}{*}{Sources} & \multicolumn{1}{c|}{\multirow{2}{*}{Method}} & \multicolumn{2}{c|}{Market-1501} & \multirow{2}{*}{Sources} & \multicolumn{1}{c|}{\multirow{2}{*}{Method}} & \multicolumn{2}{c}{DukeMTMC-reID} \\ 
\cline{3-4} \cline{7-8} & 
\multicolumn{1}{c|}{} & Rank-1 & mAP & & \multicolumn{1}{c|}{} & 
\multicolumn{1}{c|}{Rank-1} & mAP \\ 
\hline
\multirow{5}{*}{C+D+MS}  & 
DualNorm$_{50}$ & \underline{78.9} & \underline{52.3} & 
\multirow{5}{*}{C+M+MS}  & 
DualNorm$_{50}$ & \multicolumn{1}{c|}{68.5} & \underline{51.7} \\
& QAConv$_{50}$ & 65.7 & 35.6 & & 
QAConv$_{50}$ & \multicolumn{1}{c|}{66.1} & 47.1   \\
& M$^3$L (ResNet-50) & 74.5 & 48.1 & & 
M$^3$L (ResNet-50) & \multicolumn{1}{c|}{ \underline{69.4} } & 50.5   \\
& M$^3$L (IBN-Net50) & 75.9 & 50.2 & & 
M$^3$L (IBN-Net50) & \multicolumn{1}{c|}{69.2} & 51.1   \\ \cline{2-4} \cline{6-8} 
& DEX (Ours) & \textbf{81.5} & \textbf{55.2} & & 
DEX (Ours) & \multicolumn{1}{c|}{\textbf{73.7}} & \textbf{55.0}   \\ 
\hline\hline
\multirow{2}{*}{Sources} & \multicolumn{1}{c|}{\multirow{2}{*}{Method}} & \multicolumn{2}{c|}{MSMT17\_V2}  & \multirow{2}{*}{Sources} & \multicolumn{1}{c|}{\multirow{2}{*}{Method}} & \multicolumn{2}{c}{CUHK-NP}       \\ 
\cline{3-4} \cline{7-8} & 
\multicolumn{1}{c|}{} & Rank-1 & mAP & & \multicolumn{1}{c|}{} & 
\multicolumn{1}{c|}{Rank-1} & mAP \\ 
\hline
\multirow{5}{*}{C+D+M} & 
DualNorm$_{50}$ & \underline{37.9} & \underline{15.4} & 
\multirow{5}{*}{D+M+MS} & 
DualNorm$_{50}$ & \multicolumn{1}{c|}{28.0} & 27.6 \\
& QAConv$_{50}$ & 24.3 & 7.5 & & 
QAConv$_{50}$ & \multicolumn{1}{c|}{23.5} & 21.0 \\
& M$^3$L (ResNet-50) & 33.0 & 12.9 & & 
M$^3$L (ResNet-50) & \multicolumn{1}{c|}{30.7} & 29.9 \\
& M$^3$L (IBN-Net50) & 36.9 & 14.7 & & 
M$^3$L (IBN-Net50) & \multicolumn{1}{c|}{ \underline{33.1} } & \underline{32.1} \\ \cline{2-4} \cline{6-8} 
& DEX (Ours) & \textbf{43.5} & \textbf{18.7} & & 
DEX (Ours) & \multicolumn{1}{c|}{ \textbf{36.7} } & \textbf{33.8} \\ 
\hline
\end{tabular}
    \end{adjustbox}
\caption{Comparison with state-of-the-art for DG Person ReID. Bold numbers denote highest scores, while underlined numbers denote second-highest. With \Soln{} augmentation, our model surpasses the state-of-the-art in all benchmarks.}
\label{tab:soa}
\end{table}
Table \ref{tab:soa} compares our proposed solution \textbf{\Soln{}} against several recent state-of-the-art (SOTA) methods such as DualNorm~\cite{DualNorm}, QAConv~\cite{QAConv} and M$^3$L~\cite{M3L}. All experiments are evaluated on the new large scale DG Person ReID benchmarks. For clarity of presentation and alignment with subsequent ablation studies, extra experiments using CUHK-NP as one of the source domains (which were added later in \cite{M3L}) are presented in Supplemental Material. Nevertheless, our proposed solution surpasses the most recent state-of-the-art method~\cite{M3L} in \emph{all} experimental settings by a significant margin.

\subsection{Comparison with State-of-the-art Methods under Old Evaluation}

While the preceding sections focused on new evaluation benchmarks set by M$^3$L \cite{M3L}, we appreciate that much of previous work on DG Person ReID based their evaluations on the old small-scale datasets VIPeR~\cite{VIPeR}, PRID~\cite{PRID}, QMUL-GRID~\cite{GRID}, and i-LIDS~\cite{i-LIDS}. For completeness, we present our evaluation of \Soln{} on these small-scale benchmarks.  Table \ref{tab:old-eval-sota} compares our \Soln{} under this small-dataset evaluation scheme against other current state-of-the-art DG Person ReID methods. Following standard evaluation methodology, we trained on multiple large-scale benchmark datasets CUHK02~\cite{CUHK02}, CUHK03~\cite{CUHK03}, CUHK-SYSU~\cite{CUHK-SYSU}, Market-1501~\cite{Market-1501} and DukeMTMC-reID~\cite{DukeMTMC-reID} and compared against other state-of-the-art methods trained under the same setting. Combining all source datasets resulted in a total of 121,765 images with 18,530 unique PIDs. Because the large number of PIDs takes up a significant GPU memory overhead for \Soln{}, we trained our model using \SolnLite{} instead, sampling 2,000 PIDs to reduce memory use and speed up training.

\begin{table}[ht]
\centering
\begin{adjustbox}{width=\textwidth}
\begin{tabular}{l|cccc|cccc|cccc|cccc}
\hline
\multicolumn{1}{c|}{\multirow{2}{*}{Method}} & \multicolumn{4}{c|}{VIPeR (V)} & \multicolumn{4}{c|}{PRID (P)} & \multicolumn{4}{c|}{GRID (G)} & \multicolumn{4}{c}{i-LIDS (L)} \\ \cline{2-17} 
\multicolumn{1}{c|}{} & R-1    & R-5   & R-10  & mAP   & R-1   & R-5   & R-10  & mAP   & R-1   & R-5   & R-10  & mAP   & R-1    & R-5   & R-10  & mAP   \\ \hline \hline
AugMining~\cite{AugMining} & 49.8   & \underline{70.8}  & 77.0  & - & 34.3  & 56.2  & 65.7  & -     & 46.6  & \underline{67.5}  & \underline{76.1}  & - & 76.3 & \underline{93.0}  & \underline{95.3}  & - \\
DDAN~\cite{DDAN} & 56.5   & 65.6  & 76.3  & 60.8  & 62.9  & 74.2  & 85.3  & 67.5  & 46.2  & 55.4  & 68.0  & 50.9  & 78.0 & 85.7  & 93.2  & 81.2  \\
DIMN~\cite{DIMN} & 51.2 & 70.2  & 76.0  & 60.1  & 39.2  & 67.0  & 76.7 & 52.0 & 29.3  & 53.3  & 65.8  & 41.1  & 70.2 & 89.7  & 94.5  & 78.4  \\
DIR-ReID~\cite{DIR-ReID} & 58.3 & 66.9 & \underline{77.3}  & 62.9  & \textbf{71.1}  & \underline{82.4}  & \underline{88.6}  & \underline{75.6}  & 47.8  & 51.1  & 70.5  & \underline{52.1}  & 74.4   & 83.1  & 90.2  & 78.6  \\
DualNorm$_{resnet}$~\cite{DualNorm} & 59.4   & -  & -  & -  & 69.6  & -  & -  & -  & 43.7  & -  & -  & - & 78.2 & -  & -  & -  \\
MMFA-AAE~\cite{MMFA-AAE} & 58.4   & -  & -  & -  & 57.2  & -  & -  & -  & 47.4  & -  & -  & - & 84.8 & -  & -  & -  \\
RaMoE~\cite{RaMoE} & 56.6   & -  & -  & \underline{64.6}  & 57.7  & -  & -  & 67.3  & 46.8  & -  & -  & 54.2 & \underline{85.0} & -  & -  & \underline{90.2}  \\
SNR~\cite{SNR} & 52.9   & -  & -  & 61.3  & 52.1  & -  & -  & 66.5  & 40.2  & -  & -  & 47.7 & 84.1 & -  & -  & 89.9  \\
BCaR~\cite{BCaR} & \textbf{65.8} & -  & -  & -  & 70.2  & -  & -  & -  & \underline{52.8}  & -  & -  & - & 81.3 & -  & -  & -  \\
\textbf{\SolnLite{} (Ours)} & \underline{65.5} & \textbf{79.2} & \textbf{83.6} & \textbf{72.0} & \underline{71.0} & \textbf{87.8}  & \textbf{92.5}  & \textbf{78.5} & \textbf{53.3} & \textbf{69.4} & \textbf{79.0} & \textbf{61.7} & \textbf{86.3} & \textbf{95.2} & \textbf{97.3} & \textbf{90.7} \\ \hline
\end{tabular}
\end{adjustbox}
\caption{Evaluation on small-scale benchmarks VIPeR, GRID, PRID and i-LIDS.}
\label{tab:old-eval-sota}
\end{table}

\SolnLite{} demonstrates good all-round performance, surpassing the state-of-the-art methods in GRID and i-LIDS for all measures and coming in first for Rank-5, Rank-10 and mAP for all benchmarks. For VIPeR and PRID, our method is second place for Rank-1, closely matching the top performers. It is interesting to note that our Rank-5, Rank-10 and mAP scores surpass other methods by a significant margin even without re-ranking. This indicates that application of \Soln{} improves feature generalization such that \textbf{all positives} in the gallery obtain a better ranking, and attests to the effectiveness of \Soln{} as a plug-and-play method.

\subsection{Ablation Study on the Effects of each Technique}
Table \ref{tab:ablation-baseline} studies the effects of instance normalization (IN), reducing probability of random erasing (RE), ISDA~\cite{ISDA}, and \Soln{} to the BagTricks~\cite{StrongBaseline} baseline. Adding IN yields the most significant benefit. Reducing the probability of RE or applying \Soln{} also yield significant generalization benefits. Combined with IN, we tested two configurations of RE, in the first removing the augmentation entirely (NoRE), and in the other reducing the probability of applying RE on a sample from 0.5 to 0.1 (RE(0.1)). A small probability of RE seems to improve model generalization slightly. Using just IN and RE(0.1) can yield very competitive performance in all benchmarks except for CUHK-NP. Adding ISDA does not always improve the baseline. With \Soln{}, performance gains are more consistent and are especially significant in the case of CUHK-NP with a \textbf{7.7\%} improvement in Rank-1 and \textbf{5\%} increase in mAP that outperforms the current state-of-the-art by a large margin.
\begin{table}[h]
\centering
\begin{adjustbox}{width=\textwidth}
\begin{tabular}{c|l|c|c|c|l|cc}
\hline
\multirow{2}{*}{Sources} & \multicolumn{1}{c|}{\multirow{2}{*}{Method}} & \multicolumn{2}{c|}{Market-1501} & \multirow{2}{*}{Sources} & \multicolumn{1}{c|}{\multirow{2}{*}{Method}} & \multicolumn{2}{c}{DukeMTMC-reID} \\ \cline{3-4} \cline{7-8} & 
\multicolumn{1}{c|}{} & Rank-1 & mAP & & \multicolumn{1}{c|}{} & 
\multicolumn{1}{c|}{Rank-1} & mAP \\ 
\hline
\multirow{8}{*}{C+D+MS}  & 
BagTricks & 71.6 & 43.3 & 
\multirow{8}{*}{C+M+MS}  & 
BagTricks & \multicolumn{1}{c|}{58.2} & 40.8   \\
& +IN (=DualNorm) & 78.9 & 52.3 & & 
+IN (=DualNorm) & \multicolumn{1}{c|}{68.5} & 51.7   \\
& +RE(0.1) & 74.0 & 45.6 & & 
+RE(0.1) & \multicolumn{1}{c|}{63.1} & 44.6   \\
& +DEX & 71.9 & 45.4 & & 
+DEX & \multicolumn{1}{c|}{66.1} & 48.1   \\
& +IN+NoRE & 80.8 & 54.0 & & 
+IN+NoRE & \multicolumn{1}{c|}{70.4} & 52.8   \\
& +IN+RE(0.1) & 81.0 & 54.3 & & 
+IN+RE(0.1) & 
\multicolumn{1}{c|}{71.0} & 53.4   \\
& +IN+RE(0.1)+ISDA & 79.6 & 53.9 & & 
+IN+RE(0.1)+ISDA & 
\multicolumn{1}{c|}{72.3} & 54.5   \\
& +IN+RE(0.1)+DEX (\textbf{Ours}) & \textbf{81.5} & \textbf{55.2} & & 
+IN+RE(0.1)+DEX \textbf{(Ours}) & 
\multicolumn{1}{c|}{\textbf{73.7}} & \textbf{55.0}   \\ 
\hline\hline
\multirow{2}{*}{Sources} & \multicolumn{1}{c|}{\multirow{2}{*}{Method}} & \multicolumn{2}{c|}{MSMT17\_V2}  & \multirow{2}{*}{Sources} & \multicolumn{1}{c|}{\multirow{2}{*}{Method}} & \multicolumn{2}{c}{CUHK-NP}       \\ \cline{3-4} \cline{7-8} & 
\multicolumn{1}{c|}{} & Rank-1 & mAP & & \multicolumn{1}{c|}{} & 
\multicolumn{1}{c|}{Rank-1} & mAP \\ 
\hline
\multirow{8}{*}{C+D+M}   & 
BagTricks & 19.4 & 6.9 & 
\multirow{8}{*}{D+M+MS}  & 
BagTricks & \multicolumn{1}{c|}{20.1} & 19.6   \\
& +IN (=DualNorm) & 37.9 & 15.4 & & 
+IN (=DualNorm) & \multicolumn{1}{c|}{28.0} & 27.6   \\
& +RE(0.1) & 23.5 & 8.5 & & 
+RE(0.1) & \multicolumn{1}{c|}{23.9} & 23.5   \\
& +DEX & 22.9 & 8.7 & & 
+DEX & \multicolumn{1}{c|}{25.9} & 25.1   \\
& +IN+NoRE & 42.0 & 17.1 & & 
+IN+NoRE & \multicolumn{1}{c|}{28.0} & 28.2   \\
& +IN+RE(0.1) & 42.4 & 17.5 & & 
+IN+RE(0.1) & \multicolumn{1}{c|}{29.0}      & 28.8   \\
& +IN+RE(0.1)+ISDA & 41.8 & 17.7 & & 
+IN+RE(0.1)+ISDA & \multicolumn{1}{c|}{32.6} & 32.4   \\
& +IN+RE(0.1)+DEX (\textbf{Ours}) & \textbf{43.5} & \textbf{18.7}& & 
+IN+RE(0.1)+DEX (\textbf{Ours}) & \multicolumn{1}{c|}{\textbf{36.7}} & \textbf{33.8}  \\ \hline
\end{tabular}
\end{adjustbox}
\caption{Ablation study comparing the effects of applying IN, reducing RE probability, applying ISDA; and applying \Soln{} on a strong baseline ResNet-50 model from BagTricks~\cite{DualNorm}}
\label{tab:ablation-baseline}
\end{table}
\vspace{-1em}

\subsection{Applying \SolnLite{} to Multi-branch Architectures}
We demonstrate general applicability of \Soln{} by applying it on multi-branch architectures that incur a large increase in memory use. For these architectures, we can limit the sample size to apply \SolnLite{} effectively. We selected two well-known multi-branch models in supervised Person ReID: PCB~\cite{PCB}, and MGN~\cite{MGN}. PCB divides the final feature tensor into six horizontal stripes, each of which becomes a local PID prediction branch. We apply \SolnLite{} each of these local branches in PCB. MGN consists of eight PID prediction branches, with three global-level features and five part-level features. We apply \SolnLite{} to the three global-level prediction branches to contrast PCB's part-level design.
\begin{table}[h]
\centering
    \begin{adjustbox}{width=\textwidth}
\begin{tabular}{c|c|l|c|c|c|c|l|cc}
\hline
\multirow{2}{*}{Sources} & \multicolumn{1}{l|}{\multirow{2}{*}{Model}} & \multicolumn{1}{c|}{\multirow{2}{*}{Method}} & \multicolumn{2}{c|}{Market-1501} & \multirow{2}{*}{Sources} & \multicolumn{1}{l|}{\multirow{2}{*}{Model}} & \multicolumn{1}{c|}{\multirow{2}{*}{Method}} & \multicolumn{2}{c}{DukeMTMC-reID} \\ \cline{4-5} \cline{9-10} & 
\multicolumn{1}{l|}{} & 
\multicolumn{1}{c|}{} & Rank-1 & mAP & & \multicolumn{1}{l|}{} & 
\multicolumn{1}{c|}{} & 
\multicolumn{1}{c|}{Rank-1} & mAP \\ 
\hline
\multirow{4}{*}{C+D+MS} 
& \multirow{2}{*}{MGN~\cite{MGN}} & 
+IN+RE(0.1) & 62.9 & 29.7 & 
\multirow{4}{*}{C+M+MS} &
\multirow{2}{*}{MGN~\cite{MGN}} & 
+IN+RE(0.1) & \multicolumn{1}{c|}{62.2} & 40.8 \\
& & +IN+RE(0.1)+\SolnLite{} & \textbf{63.7} & \textbf{32.8} & & &
+IN+RE(0.1)+\SolnLite{} & \multicolumn{1}{c|}{\textbf{64.3}} & \textbf{44.4} \\ \cline{2-5} \cline{7-10} & 
\multirow{2}{*}{PCB~\cite{PCB}} & 
+IN+RE(0.1) & 71.2 & 42.4 & & 
\multirow{2}{*}{PCB~\cite{PCB}} & 
+IN+RE(0.1) & \multicolumn{1}{c|}{65.5} & 45.1 \\
& & +IN+RE(0.1)+\SolnLite{} & \textbf{73.1} & \textbf{45.1} & & & 
+IN+RE(0.1)+\SolnLite{} & \multicolumn{1}{c|}{\textbf{66.5}} & \textbf{46.8} \\ \hline\hline
\multirow{2}{*}{Sources} & \multicolumn{1}{l|}{\multirow{2}{*}{Model}} & \multicolumn{1}{c|}{\multirow{2}{*}{Method}} & \multicolumn{2}{c|}{MSMT17\_V2}  & \multirow{2}{*}{Sources} & \multicolumn{1}{l|}{\multirow{2}{*}{Model}} & \multicolumn{1}{c|}{\multirow{2}{*}{Method}} & \multicolumn{2}{c}{CUHK-NP} \\ 
\cline{4-5} \cline{9-10} & 
\multicolumn{1}{l|}{} & 
\multicolumn{1}{c|}{} & Rank-1 & mAP & & 
\multicolumn{1}{l|}{} & 
\multicolumn{1}{c|}{} & 
\multicolumn{1}{c|}{Rank-1} & mAP \\ 
\hline
\multirow{4}{*}{C+D+M}   & 
\multirow{2}{*}{MGN~\cite{MGN}} & 
+IN+RE(0.1) & 23.1 & 8.3 & 
\multirow{4}{*}{D+M+MS}  & 
\multirow{2}{*}{MGN~\cite{MGN}} & 
+IN+RE(0.1) & \multicolumn{1}{c|}{18.1} & 15.5 \\
& & +IN+RE(0.1)+\SolnLite{} & \textbf{25.9} & \textbf{9.6} & & &
+IN+RE(0.1)+\SolnLite{} & \multicolumn{1}{c|}{\textbf{22.9}} & \textbf{21.0} \\ 
\cline{2-5} \cline{7-10} & 
\multirow{2}{*}{PCB~\cite{PCB}} & 
+IN+RE(0.1) & 36.4 & 14.6 & & 
\multirow{2}{*}{PCB~\cite{PCB}} & 
+IN+RE(0.1) & \multicolumn{1}{c|}{25.2} & 25.2 \\
& & +IN+RE(0.1)+\SolnLite{} & \textbf{37.3} & \textbf{15.2} & & &
+IN+RE(0.1)+\SolnLite{} & 
\multicolumn{1}{c|}{\textbf{28.4}} & \textbf{27.2} \\ \hline
\end{tabular} 
\end{adjustbox}
\caption{Applying \SolnLite{} to multi-branch architecture models such MGN~\cite{MGN} and PCB~\cite{PCB} similarly yield performance improvements}
\label{tab:ablation-mgnpcb}
\end{table}

Figure \ref{fig:our-method}(b) illustrates the multi-branch application of our technique using PCB as an example. MGN's loss is averaged among all branches, so we kept $\lambda$ at 7.5, same as our original baseline experiments. However, PCB sums the losses from all branches instead of averaging so we reduced $\lambda$ by a factor of six (the number of stripes in PCB) to 1.25. We set the sample size to 2,000 for \SolnLite{} and train for 40 epochs. Table \ref{tab:ablation-mgnpcb} shows the result of our comparison study. Overall, \Soln{} demonstrates its effectiveness as an augmentation strategy and ability to improve the generalization results of a range of models.

\subsection{Ablation Study on Negative Sample Sizes for \SolnLite{}}
We investigate the effects of different sample sizes when applying negative sampling in \SolnLite{}, shown in Table \ref{tab:ablation-negsample}. The performance improves as the sample size increases but starts to plateau beyond a sample size of around 2,000. With 2,000 negative sample sizes for estimation, \SolnLite{} can achieve comparable performance on par with the full \Soln{} with much faster training speed and memory usage.

\begin{table}[h]
\centering
\begin{adjustbox}{width=\textwidth}
\begin{tabular}{c|c|c|c|c|c|c|c|c|c}
\hline
\multicolumn{1}{l|}{\multirow{2}{*}{Sources}} & \multirow{2}{*}{Type} & \multirow{2}{*}{Samples} & \multicolumn{2}{c|}{Market-1501} & \multicolumn{1}{l|}{\multirow{2}{*}{Sources}} & \multirow{2}{*}{Type} & \multirow{2}{*}{Samples} & \multicolumn{2}{c}{DukeMTMC-reID} \\ \cline{4-5} \cline{9-10} 
\multicolumn{1}{l|}{} &  &  & Rank-1 & mAP & \multicolumn{1}{l|}{} &  &  & Rank-1 & mAP \\ \hline
\multirow{5}{*}{C+D+MS} & \multirow{4}{*}{DEXLite} & 10 & 54.6 & 30.0 & \multirow{5}{*}{C+M+MS} & \multirow{4}{*}{DEXLite} & 10 & 53.2 & 34.2 \\
 &  & 100 & 69.7 & 42.8 &  &  & 100 & 64.8 & 46.0 \\
 &  & 1000 & 77.0 & 51.4 &  &  & 1000 & 70.8 & 52.6 \\
 &  & 2000 & 80.3 & 54.3 &  &  & 2000 & 72.1 & 53.6 \\ \cline{2-5} \cline{7-10} 
 & DEX & Full & 81.5 & 55.2 &  & DEX & Full & 72.7 & 54.2 \\ \hline \hline
\multicolumn{1}{l|}{\multirow{2}{*}{Sources}} & \multirow{2}{*}{Type} & \multirow{2}{*}{Sample} & \multicolumn{2}{c|}{MSMT17\_V2} & \multicolumn{1}{l|}{\multirow{2}{*}{Sources}} & \multirow{2}{*}{Type} & \multirow{2}{*}{Samples} & \multicolumn{2}{c}{CUHK-NP} \\ \cline{4-5} \cline{9-10} 
\multicolumn{1}{l|}{} &  &  & \multicolumn{1}{l|}{Rank-1} & \multicolumn{1}{l|}{mAP} & \multicolumn{1}{l|}{} &  &  & \multicolumn{1}{l|}{Rank-1} & \multicolumn{1}{l}{mAP} \\ \hline
\multirow{5}{*}{D+M+MS} & \multirow{4}{*}{DEXLite} & 10 & 16.8 & 16.8 & \multirow{5}{*}{C+D+M} & \multirow{4}{*}{DEXLite} & 10 & 17.5 & 6.5 \\
 &  & 100 & 27.6 & 27.0 &  &  & 100 & 31.3 & 12.7 \\
 &  & 1000 & 32.8 & 32.1 &  &  & 1000 & 38.6 & 16.4 \\
 &  & 2000 & 34.4 & 33.3 &  &  & 2000 & 41.5 & 17.6 \\ \cline{2-5} \cline{7-10} 
 & DEX & Full & 34.3 & 32.9 &  & DEX & Full & 43.5 & 18.7 \\ \hline
\end{tabular}
\end{adjustbox}
\caption{Effects of sample sizes in \SolnLite{} in comparison with \Soln{}}
\label{tab:ablation-negsample}
\end{table}

\section{Conclusion}
\label{sec:conclusion}

In this paper, we introduced a fresh perspective and a novel solution to the problem of domain generalization (DG) in Person ReID, leveraging on domain biases inherently encoded in deep features to augment them directly in domain-semantically meaningful directions. \Soln{}, our proposed dual-label implicit augmentation method applied in simple ResNet50 network can surpasses the most of recent state-of-the-art Domain Generalization (DG) Person ReID methods in both new large-scale banchmarks and old small-scale benchmarks by a large margins. Reducing the memory use of our method with negative sampling techniques, we also developed \SolnLite{} to make our method applicable to a wide range of model architectures and dataset situations. There is still room to grow before DG Person ReID solutions are robust enough for true real-world use, and we believe that our novel approach will seed more developments along this new paradigm and pave the way for faster innovation.\\

\noindent
\textbf{Acknowledgement}: This work was supported by the Defence Science and Technology Agency (DSTA) Postgraduate Scholarship, of which Eugene P.W. Ang is a recipient. It was carried out at the Rapid-Rich Object Search (ROSE) Lab at the Nanyang Technological University, Singapore.


\bibliography{lsbib}

\begin{thebibliography}{39}
\providecommand{\natexlab}[1]{#1}
\providecommand{\url}[1]{\texttt{#1}}
\expandafter\ifx\csname urlstyle\endcsname\relax
  \providecommand{\doi}[1]{doi: #1}\else
  \providecommand{\doi}{doi: \begingroup \urlstyle{rm}\Url}\fi

\bibitem[Chen et~al.(2020)Chen, Dai, Liu, Zheng, Tian, and Ji]{DDAN}
Peixian Chen, Pingyang Dai, Jianzhuang Liu, Feng Zheng, Qi~Tian, and Rongrong
  Ji.
\newblock {Dual Distribution Alignment Network for Generalizable Person
  Re-Identification}.
\newblock In \emph{Proc. AAAI Conference on Artificial Intelligence (AAAI)},
  2020.

\bibitem[{Chen Change Loy} et~al.(2009){Chen Change Loy}, {Tao Xiang}, and
  {Shaogang Gong}]{GRID}
{Chen Change Loy}, {Tao Xiang}, and {Shaogang Gong}.
\newblock {Multi-camera activity correlation analysis}.
\newblock In \emph{Proc. IEEE Conference on Computer Vision and Pattern
  Recognition (CVPR)}, pages 1988--1995, 2009.

\bibitem[Dai et~al.(2021)Dai, Li, Liu, Tong, and Duan]{RaMoE}
Yongxing Dai, Xiaotong Li, Jun Liu, Zekun Tong, and Ling-Yu Duan.
\newblock {Generalizable Person Re-identification with Relevance-aware Mixture
  of Experts}.
\newblock In \emph{Proc. IEEE Conference on Computer Vision and Pattern
  Recognition (CVPR)}, 2021.

\bibitem[Deng et~al.(2018)Deng, Zheng, Ye, Kang, Yang, and Jiao]{SPGAN}
Weijian Deng, Liang Zheng, Qixiang Ye, Guoliang Kang, Yi~Yang, and Jianbin
  Jiao.
\newblock {Image-Image Domain Adaptation with Preserved Self-Similarity and
  Domain-Dissimilarity for Person Re-identification}.
\newblock In \emph{Proc. IEEE Conference on Computer Vision and Pattern
  Recognition (CVPR)}, pages 994--1003, 2018.

\bibitem[Goodfellow et~al.(2014)Goodfellow, Pouget-Abadie, Mirza, Xu,
  Warde-Farley, Ozair, Courville, and Bengio]{Goodfellow2014GenerativeNets}
Ian~J. Goodfellow, Jean Pouget-Abadie, Mehdi Mirza, Bing Xu, David
  Warde-Farley, Sherjil Ozair, Aaron Courville, and Yoshua Bengio.
\newblock {Generative Adversarial Nets}.
\newblock In \emph{Proc. Advances in Neural Information Processing Systems
  (NeurIPS)}, page 2672–2680, 2014.

\bibitem[Gray et~al.(2007)Gray, Brennan, and Tao]{VIPeR}
Doug Gray, Shane Brennan, and Hai Tao.
\newblock {Evaluating Appearance Models for Recognition, Reacquisition, and
  Tracking}.
\newblock In \emph{International Workshop on Performance Evaluation for
  Tracking and Surveillance (PETS)}, 2007.

\bibitem[He et~al.(2016)He, Zhang, Ren, and Sun]{ResNet}
Kaiming He, Xiangyu Zhang, Shaoqing Ren, and Jian Sun.
\newblock {Deep Residual Learning for Image Recognition}.
\newblock In \emph{Proc. IEEE Conference on Computer Vision and Pattern
  Recognition (CVPR)}, pages 770--778, 2016.

\bibitem[Hendrycks et~al.(2020)Hendrycks, Basart, Mu, Kadavath, Wang, Dorundo,
  Desai, Zhu, Parajuli, Guo, Song, Steinhardt, and Gilmer]{DeepAugment}
Dan Hendrycks, Steven Basart, Norman Mu, Saurav Kadavath, Frank Wang, Evan
  Dorundo, Rahul Desai, Tyler Zhu, Samyak Parajuli, Mike Guo, Dawn Song, Jacob
  Steinhardt, and Justin Gilmer.
\newblock {The Many Faces of Robustness: A Critical Analysis of
  Out-of-Distribution Generalization}.
\newblock \emph{arXiv preprint}, 2020.

\bibitem[Hermans et~al.(2017)Hermans, Beyer, and Leibe]{TriNet}
Alexander Hermans, Lucas Beyer, and Bastian Leibe.
\newblock {In Defense of the Triplet Loss for Person Re-Identification}.
\newblock In \emph{arXiv preprint}, 3 2017.

\bibitem[Hirzer et~al.(2011)Hirzer, Beleznai, Roth, and Bischof]{PRID}
Martin Hirzer, Csaba Beleznai, Peter~M Roth, and Horst Bischof.
\newblock {Person Re-identification by Descriptive and Discriminative
  Classification}.
\newblock In \emph{Scandinavian Conference on Image Analysis (SCIA)}, 2011.

\bibitem[Jia et~al.(2019)Jia, Ruan, and Hospedales]{DualNorm}
Jieru Jia, Qiuqi Ruan, and Timothy~M. Hospedales.
\newblock {Frustratingly Easy Person Re-Identification: Generalizing Person
  Re-ID in Practice}.
\newblock In \emph{Proc. British Machine Vision Conference (BMVC)}, 2019.

\bibitem[Jin et~al.(2020)Jin, Lan, Zeng, Chen, and Zhang]{SNR}
Xin Jin, Cuiling Lan, Wenjun Zeng, Zhibo Chen, and Li~Zhang.
\newblock {Style Normalization and Restitution for Generalizable Person
  Re-identification}.
\newblock In \emph{Proc. IEEE Conference on Computer Vision and Pattern
  Recognition (CVPR)}, 2020.

\bibitem[Kingma and Ba(2015)]{Kingma2015Adam:Optimization}
Diederik~P. Kingma and Jimmy Ba.
\newblock {Adam: A Method for Stochastic Optimization}.
\newblock In \emph{Proc. International Conference on Learning Representations
  (ICLR)}, 2015.

\bibitem[Li and Wang(2013)]{CUHK02}
Wei Li and Xiaogang Wang.
\newblock {Locally Aligned Feature Transforms across Views}.
\newblock In \emph{Proc. IEEE Conference on Computer Vision and Pattern
  Recognition (CVPR)}, pages 3594--3601, 2013.

\bibitem[Liao and Shao(2020)]{QAConv}
Shengcai Liao and Ling Shao.
\newblock {Interpretable and Generalizable Person Re-identification with
  Query-Adaptive Convolution and Temporal Lifting}.
\newblock In \emph{Proc. European Conference on Computer Vision (ECCV)}, 2020.

\bibitem[Lin et~al.(2018)Lin, Li, Li, and Kot]{MMFA}
Shan Lin, Haoliang Li, Chang-Tsun Li, and Alex~C Kot.
\newblock {Multi-task Mid-level Feature Alignment Network for Unsupervised
  Cross-Dataset Person Re-Identification}.
\newblock In \emph{Proc. British Machine Vision Conference (BMVC)}, 2018.

\bibitem[Lin et~al.(2021)Lin, Li, and Kot]{MMFA-AAE}
Shan Lin, Chang-tsun Li, and Alex~C Kot.
\newblock {Multi-Domain Adversarial Feature Generalization for Person
  Re-Identification}.
\newblock \emph{IEEE Transactions on Image Processing}, 2021.

\bibitem[Luo et~al.(2019)Luo, Gu, Liao, Lai, and Jiang]{StrongBaseline}
Hao Luo, Youzhi Gu, Xingyu Liao, Shenqi Lai, and Wei Jiang.
\newblock {Bag of Tricks and a Strong Baseline for Deep Person
  Re-Identification}.
\newblock In \emph{Proc. IEEE Conference on Computer Vision and Pattern
  Recognition Workshops (CVPRW)}, pages 1487--1495, 2019.

\bibitem[Mikolov et~al.(2013)Mikolov, Chen, Corrado, and Dean]{Word2Vec}
Tomas Mikolov, Kai Chen, Greg Corrado, and Jeffrey Dean.
\newblock {Efficient estimation of word representations in vector space}.
\newblock In \emph{Proc. International Conference on Learning Representations
  Workshop (ICLRW)}, pages 1--12, 2013.

\bibitem[Peng et~al.(2016)Peng, Xiang, Wang, Pontil, Gong, Huang, and
  Tian]{UMDL}
Peixi Peng, Tao Xiang, Yaowei Wang, Massimiliano Pontil, Shaogang Gong, Tiejun
  Huang, and Yonghong Tian.
\newblock {Unsupervised Cross-Dataset Transfer Learning for Person
  Re-identification}.
\newblock In \emph{Proc. IEEE Conference on Computer Vision and Pattern
  Recognition (CVPR)}, pages 1306--1315, 2016.

\bibitem[Qiao et~al.(2018)Qiao, Liu, Shen, and Yuille]{CUHK-SYSU}
Siyuan Qiao, Chenxi Liu, Wei Shen, and Alan Yuille.
\newblock {Few-Shot Image Recognition by Predicting Parameters from
  Activations}.
\newblock In \emph{Proc. IEEE Conference on Computer Vision and Pattern
  Recognition (CVPR)}, pages 7229--7238, 2018.
\newblock ISBN 978-1-5386-6420-9.
\newblock \doi{10.1109/CVPR.2018.00755}.

\bibitem[Song et~al.(2019)Song, Yang, Song, Xiang, and Hospedales]{DIMN}
Jifei Song, Yongxin Yang, Yi-Zhe Song, Tao Xiang, and Timothy~M. Hospedales.
\newblock {Generalizable Person Re-Identification by Domain-Invariant Mapping
  Network}.
\newblock In \emph{Proc. IEEE Conference on Computer Vision and Pattern
  Recognition (CVPR)}, 2019.

\bibitem[Sun et~al.(2018)Sun, Zheng, Yang, Tian, and Wang]{PCB}
Yifan Sun, Liang Zheng, Yi~Yang, Qi~Tian, and Shengjin Wang.
\newblock {Beyond Part Models: Person Retrieval with Refined Part Pooling}.
\newblock In \emph{Proc. European Conference on Computer Vision (ECCV)}, 2018.

\bibitem[Tamura and Murakami(2019)]{AugMining}
Masato Tamura and Tomokazu Murakami.
\newblock {Augmented Hard Example Mining for Generalizable Person
  Re-Identification}.
\newblock In \emph{arXiv preprint}, 2019.

\bibitem[Tamura and Yoshinaga(2020)]{BCaR}
Masato Tamura and Tomoaki Yoshinaga.
\newblock {BCaR : Beginner Classifier as Regularization Towards Generalizable
  Re-ID}.
\newblock In \emph{Proc. British Machine Vision Conference (BMVC)}, 2020.

\bibitem[Upchurch et~al.(2017)Upchurch, Gardner, Pleiss, Pless, Snavely, Bala,
  and Weinberger]{DFI}
Paul Upchurch, Jacob Gardner, Geoff Pleiss, Robert Pless, Noah Snavely, Kavita
  Bala, and Kilian Weinberger.
\newblock {Deep feature interpolation for image content changes}.
\newblock In \emph{Proc. IEEE Conference on Computer Vision and Pattern
  Recognition (CVPR)}, pages 6090--6099, 2017.

\bibitem[Wang et~al.(2016)Wang, Zuo, Lin, Zhang, and Zhang]{CUHK03}
Faqiang Wang, Wangmeng Zuo, Liang Lin, David Zhang, and Lei Zhang.
\newblock {Joint Learning of Single-Image and Cross-Image Representations for
  Person Re-identification}.
\newblock In \emph{Proc. IEEE Conference on Computer Vision and Pattern
  Recognition (CVPR)}, pages 1288--1296, 2016.

\bibitem[Wang et~al.(2018{\natexlab{a}})Wang, Yuan, Chen, Li, and Zhou]{MGN}
Guanshuo Wang, Yufeng Yuan, Xiong Chen, Jiwei Li, and Xi~Zhou.
\newblock {Learning Discriminative Features with Multiple Granularities for
  Person Re-Identification}.
\newblock In \emph{Proc. ACM International Conference on Multimedia (ACM MM)},
  pages 274--282, 2018{\natexlab{a}}.

\bibitem[Wang et~al.(2018{\natexlab{b}})Wang, Zhu, Gong, and Li]{TJ-AIDL}
Jingya Wang, Xiatian Zhu, Shaogang Gong, and Wei Li.
\newblock {Transferable Joint Attribute-Identity Deep Learning for Unsupervised
  Person Re-identification}.
\newblock In \emph{Proc. IEEE Conference on Computer Vision and Pattern
  Recognition (CVPR)}, pages 2275--2284, 2018{\natexlab{b}}.

\bibitem[Wang et~al.(2019)Wang, Pan, Song, Zhang, Wu, and Huang]{ISDA}
Yulin Wang, Xuran Pan, Shiji Song, Hong Zhang, Cheng Wu, and Gao Huang.
\newblock {Implicit Semantic Data Augmentation for Deep Networks}.
\newblock In \emph{Proc. Advances in Neural Information Processing Systems
  (NeurIPS)}, 2019.

\bibitem[Wei et~al.(2018)Wei, Zhang, Gao, and Tian]{MSMT17}
Longhui Wei, Shiliang Zhang, Wen Gao, and Qi~Tian.
\newblock {Person Transfer GAN to Bridge Domain Gap for Person
  Re-identification}.
\newblock In \emph{Proc. IEEE Conference on Computer Vision and Pattern
  Recognition (CVPR)}, pages 79--88, 2018.

\bibitem[Wen et~al.(2016)Wen, Zhang, Li, and Qiao]{Wen2016ARecognition}
Yandong Wen, Kaipeng Zhang, Zhifeng Li, and Yu~Qiao.
\newblock {A Discriminative Feature Learning Approach for Deep Face
  Recognition}.
\newblock In \emph{Proc. European Conference on Computer Vision (ECCV)}, pages
  499--515, 2016.

\bibitem[Zhang et~al.(2021)Zhang, Zhang, Zhang, Li, Jia, Wang, and
  Tan]{DIR-ReID}
Yi-Fan Zhang, Hanlin Zhang, Zhang Zhang, Da~Li, Zhen Jia, Liang Wang, and
  Tieniu Tan.
\newblock {Learning Domain Invariant Representations for Generalizable Person
  Re-Identification}.
\newblock \emph{arXiv preprint}, 2021.

\bibitem[Zhao et~al.(2021)Zhao, Zhong, Yang, Luo, Lin, Li, and Sebe]{M3L}
Yuyang Zhao, Zhun Zhong, Fengxiang Yang, Zhiming Luo, Yaojin Lin, Shaozi Li,
  and Nicu Sebe.
\newblock {Learning to Generalize Unseen Domains via Memory-based Multi-Source
  Meta-Learning for Person Re-Identification}.
\newblock In \emph{Proc. IEEE Conference on Computer Vision and Pattern
  Recognition (CVPR)}, 2021.

\bibitem[Zheng et~al.(2015)Zheng, Shen, Tian, Wang, Wang, and
  Tian]{Market-1501}
Liang Zheng, Liyue Shen, Lu~Tian, Shengjin Wang, Jingdong Wang, and Qi~Tian.
\newblock {Scalable Person Re-identification: A Benchmark}.
\newblock In \emph{Proc. IEEE International Conference on Computer Vision
  (ICCV)}, pages 1116--1124, 2015.

\bibitem[Zheng et~al.(2009)Zheng, Gong, and Xiang]{i-LIDS}
Wei~Shi Zheng, Shaogang Gong, and Tao Xiang.
\newblock {Associating Groups of People}.
\newblock In \emph{Proc. British Machine Vision Conference (BMVC)}, pages
  23--1, 2009.
\newblock ISBN 1901725391.
\newblock \doi{10.5244/C.23.23}.

\bibitem[Zheng et~al.(2017)Zheng, Zheng, and Yang]{DukeMTMC-reID}
Zhedong Zheng, Liang Zheng, and Yi~Yang.
\newblock {Unlabeled Samples Generated by GAN Improve the Person
  Re-identification Baseline in Vitro}.
\newblock In \emph{Proc. IEEE International Conference on Computer Vision
  (ICCV)}, pages 3774--3782, 2017.

\bibitem[Zheng et~al.(2019)Zheng, Yang, Yu, Zheng, Yang, and Kautz]{DG-Net}
Zhedong Zheng, Xiaodong Yang, Zhiding Yu, Liang Zheng, Yi~Yang, and Jan Kautz.
\newblock {Joint Discriminative and Generative Learning for Person
  Re-Identification}.
\newblock In \emph{Proc. IEEE Conference on Computer Vision and Pattern
  Recognition (CVPR)}, pages 2133--2142, 2019.

\bibitem[Zhong et~al.(2017)Zhong, Zheng, Cao, and Li]{Re-ranking}
Zhun Zhong, Liang Zheng, Donglin Cao, and Shaozi Li.
\newblock {Re-ranking person re-identification with k-reciprocal encoding}.
\newblock In \emph{Proc. IEEE Conference on Computer Vision and Pattern
  Recognition (CVPR)}, pages 3652--3661, 2017.

\end{thebibliography}
\end{document}


\title{Supplemental Material}  

\maketitle
\thispagestyle{empty}

\section{Comparison with M$^3$L with C-NP as Source Domain}
\label{sec:sota-c-np}
To ensure a comprehensive comparison with ~\cite{M3L}, we present additional experimental results using the new protocol of CUHK03~\cite{CUHK03}, C-NP, as a source domain, which were omitted from the main section due to space constraints. Table \ref{tab:src-tgt-dataset-setup-c-np} summarizes each new experiment, detailing the source domains used, target domain and ID/image counts for each.
\begin{table}[ht]
\centering
    \begin{tabular}{l|l|c|c|c}
        \hline
        Sources & Target & Combined-Src-IDs & Combined-Src-Images & Target-Images \\
        \hline\hline
        C-NP+D+MS & M & 2,510 & 56,508 & 19,281 \\
        C-NP+M+MS & D & 2,559 & 52,922 & 19,889 \\
        C-NP+D+M & MS & 2,220 & 36,823 & 93,820 \\
        \hline
    \end{tabular}
    \caption{Details for experiments involving C-NP in Sources.}
    \label{tab:src-tgt-dataset-setup-c-np}
\end{table}

\noindent Table \ref{tab:soa-c-np} compares our \Soln{} with~\cite{M3L}. Under this different experimental setting, our method continues to surpass leading state-of-the-art methods by significant margins.

\begin{table}[h]
\centering
\begin{tabular}{c|l|cc}
\hline
\multirow{2}{*}{Sources}   & \multicolumn{1}{c|}{\multirow{2}{*}{Method}} & \multicolumn{2}{c}{Market-1501} \\ 
\cline{3-4} & 
\multicolumn{1}{c|}{} & 
\multicolumn{1}{c|}{Rank-1} & mAP \\
\hline
\multirow{5}{*}{C-NP+D+MS} & 
DualNorm$_{50}$ & \multicolumn{1}{c|}{76.5} & 48.5 \\
& QAConv$_{50}$ & \multicolumn{1}{c|}{68.6} & 39.5 \\
& M$^3$L (ResNet-50) & \multicolumn{1}{c|}{76.5} & 51.1 \\
& M$^3$L (IBN-Net50) & \multicolumn{1}{c|}{78.3} & 52.5 \\ \cline{2-4} & 
DEX (Ours) & \multicolumn{1}{c|}{ \textbf{79.7}  } & \textbf{53.3}  \\
\hline\hline
\multirow{2}{*}{Sources} & \multicolumn{1}{c|}{\multirow{2}{*}{Method}} & \multicolumn{2}{c}{DukeMTMC-reID} \\ 
\cline{3-4} & 
\multicolumn{1}{c|}{} & 
\multicolumn{1}{c|}{Rank-1} & mAP \\
\hline
\multirow{5}{*}{C-NP+M+MS} &
DualNorm$_{50}$ & \multicolumn{1}{c|}{66.1} & 48.8 \\
& QAConv$_{50}$ & \multicolumn{1}{c|}{64.9} & 43.4 \\
& M$^3$L (ResNet-50) & \multicolumn{1}{c|}{67.1} & 48.2 \\
& M$^3$L (IBN-Net50) & \multicolumn{1}{c|}{67.2} & 48.8 \\ \cline{2-4} & 
DEX (Ours) & \multicolumn{1}{c|}{ \textbf{72.1} } & \textbf{53.5} \\
\hline\hline
\multirow{2}{*}{Sources} & \multicolumn{1}{c|}{\multirow{2}{*}{Method}} & \multicolumn{2}{c}{MSMT17\_V2} \\
\cline{3-4} & 
\multicolumn{1}{c|}{} & 
\multicolumn{1}{c|}{Rank-1} & mAP \\
\hline
\multirow{5}{*}{C-NP+D+M} &
DualNorm$_{50}$ & \multicolumn{1}{c|}{34.4} & 13.5 \\
& QAConv$_{50}$ & \multicolumn{1}{c|}{29.9} & 10.0 \\
& M$^3$L (ResNet-50) & \multicolumn{1}{c|}{32.0} & 13.1 \\
& M$^3$L (IBN-Net50) & \multicolumn{1}{c|}{37.1} & 15.4 \\ \cline{2-4} & 
DEX (Ours) & \multicolumn{1}{c|}{ \textbf{42.7} } & \textbf{17.9} \\ 
\hline
\end{tabular}
\caption{Comparison of our \Soln{} against the most recent state-of-the-art DG Person ReID, for the experiments that use CUHK-NP in Sources.}
\label{tab:soa-c-np}
\end{table}

\section{Detailed Differences between \Soln{} and ISDA}
\label{sec:diff-datasets-isda-reid}

Our method is inspired by ISDA~\cite{ISDA}, a recent work on deep feature augmentation. Still, it is significantly different as we found it impractical to apply it, originally proposed for image classification tasks, directly to multi-source domain generalization for Person ReID (DG-ReID). In the original formulation, a different covariance matrix is stored and updated for each class corresponding to person identities in ReID. However, in DG-ReID the number of classes grows large when multiple source datasets are merged for training, imposing a strong memory overhead as each matrix takes up $O(n^2)$ space with feature dimension $n=2048$ in our case. In our case, the experiment with the lowest number of training classes is already close to 2,500, making direct application of~\cite{ISDA} infeasible. 

For problems with a large number of classes such as ImageNet~\cite{ImageNet}, ISDA approximated the class-conditional covariance by storing just the matrix diagonals, reducing the per-class memory overhead to $O(n)$ while still reaping the benefits. However, even with this approximation, using a class-conditional approach could not consistently outperform our baseline model. Looking deeper, we believe this is because DG-ReID datasets have far fewer samples per class compared to those used in the original work, making estimation of per-class covariance matrices unstable. Datasets used in~\cite{ISDA} have 500 samples per class in the case of CIFAR-100 and 5000 samples per class in CIFAR-10~\cite{CIFAR}. In ImageNet~\cite{ImageNet}, a majority of the classes have 1300 samples each, with the lowest being 732. In stark contrast, the median samples-per-identity is between 10 and 25 for ReID, \emph{over an order of magnitude smaller}. Table \ref{tab:sup-samples-per-pid} shows a detailed comparison between the dataset statistics of their datasets and our ReID datasets. Such drastically different distributions indicate that class-conditional covariances may not be the most stable solution. 


\begin{table}[ht]
\centering
    \begin{tabular}{l|c|c|c|c}
        \hline
        Dataset(s) & Mean & Median & Min & Max \\
        \hline \hline
        CIFAR-10 & 5000 & 5000 & 5000 & 5000 \\
        CIFAR-100 & 500 & 500 & 500 & 500 \\
        ImageNet~\cite{ImageNet} & 1281.2 & 1300 & 732 & 1300 \\
        \hline \hline
        Market-1501 (M) & 17.2 & 15 & 2 & 72 \\
        DukeMTMC-reID (D) & 23.5 & 20 & 6 & 426 \\
        CUHK03 (C) & 9.6 & 10 & 6 & 10 \\
        MSMT17\_V2 (MS) & 30.8 & 25 & 6 & 392 \\
        \hline \hline
        C+D+MS & 20.0 & 13 & 6 & 426 \\
        C+D+M  & 15.1 & 10 & 2 & 426 \\
        C+M+MS & 18.6 & 10 & 2 & 392 \\
        D+M+MS & 24.9 & 20 & 2 & 426 \\
        \hline \hline
    \end{tabular}
    \caption{Comparison of samples per class between the datasets used in previous work on deep feature augmentation~\cite{ISDA} and those from Person ReID. ReID datasets have far fewer samples per class, making estimation of class-conditional covariance matrices unstable.}
    \label{tab:sup-samples-per-pid}
\end{table}

\section{Model Training Details}
\label{sec:baseline-training-details}

In this section we describe the full details of our training method.

\subsection{Backbone model} We use an ImageNet~\cite{ImageNet} pre-trained ResNet-50~\cite{ResNet} with instance normalization applied on the first three out of four Bottleneck blocks, as prescribed in DualNorm~\cite{DualNorm}. The final layer stride is reduced to 1 (originally 2). A batch normalization layer with no bias is applied to the final layer features~\cite{DualNorm}.

\subsection{Loss functions} 
For regularization, the softmax loss $\Lagr_{soft}$ applies label smoothing with $\epsilon=0.1$.
The triplet loss is defined as such: \[\Lagr_{tri} = \sum_{a,p,n} [\delta(\mathbf{x_a},\mathbf{x_n}) - \delta(\mathbf{x_a},\mathbf{x_p}) + \gamma]_{+} \]where $[\cdot]_{+} = max(\cdot, 0)$, ($\mathbf{x_a},\mathbf{x_p},\mathbf{x_n}$) refer to the triplets (anchor, positive, negative) that are found in the batch, $\gamma=0.3$ is the triplet margin, and $\delta$ is a metric; in our case, we use Euclidean distance. Our batch sampling scheme follows~\cite{StrongBaseline}, selecting $k=4$ samples per PID in the batch to ensure sufficient triplets for training. The center loss is defined as such: \[ \Lagr_{cen} = \sum_{i=1}^{m} \lVert \mathbf{x_i} - \mathbf{c_{y_i}} \rVert^2 \]where $i$ indexes a batch size of $m$, $y_i$ is the class label for $x_i$ and $c$ are the class centroids, updated dynamically during training. To recap, the overall loss function for training a baseline model without \Soln{} is: \[ \Lagr_{base} = \beta_{soft} \Lagr_{soft} + \beta_{tri} \Lagr_{tri} + \beta_{cen} \Lagr_{cen} \]with $\beta_{soft}=1,\beta_{tri}=1,\beta_{cen}=5\times10^{-4}$. Applying \Soln{} or \SolnLite{}, the overall loss then changes to: \[\Lagr_{overall}(t) = \Lagr_{\Soln{}/\SolnLite{}}(\lambda_{t}) + \beta_{tri} \Lagr_{tri} + \beta_{cen} \Lagr_{cen}\]with the strength of augmentation tempered by epoch $t$ as described in the main paper.

\subsection{Input pre-processing}

We resized the input to [384,128], pad it 10 pixels around with zeros and crop out [384,128]. We then apply a horizontal flip to the image with probability $p=0.5$ and random-erasing (RE) augmentation with a probability $p=0.1$.

\subsection{Other training details} 
We train for 60 epochs ($T=60$) with a batch size of 32 using the Adam optimizer, linearly warming up the learning rate $\eta$ from 0 to $1.75 \times 10^{-4}$ in 10 epochs. Afterwards, $\eta$ is reduced by a factor of 0.1 at epochs 30 and 55.

\bibliography{lsbib}